\documentclass[12pt]{article}

\usepackage[bitstream-charter]{mathdesign}
\usepackage[mathcal]{eucal}

\usepackage{graphicx}
\usepackage[bottom]{footmisc}
\usepackage{url}
\usepackage{subfigure}

\usepackage[cmex10]{amsmath}
\usepackage{amsfonts}

\usepackage{algorithmic}
\usepackage{setspace}

\usepackage{url}
\usepackage[table]{xcolor}
\usepackage{multirow}
\usepackage{booktabs}
\usepackage{flushend}
\usepackage[colorlinks=true]{hyperref}
\usepackage{a4wide}

\usepackage{fancyhdr}
\fancypagestyle{plain}{%
	\fancyhf{}
	\fancyhead[R]{Paper published in \textbf{Cognitive Computation} (editorial preprint)}
}

\newtheorem{theorem}{Theorem}

\DeclareMathSymbol{\R}{\mathalpha}{AMSb}{"52}

\usepackage{booktabs}
\usepackage{bm}

\definecolor{LightGray}{gray}{0.85}

\title{Why should we add early exits \\ to neural networks?}

\author{Simone Scardapane, Michele Scarpiniti, Enzo Baccarelli, Aurelio Uncini \thanks{All authors are with the Dept. of Information Engineering, Electronics and Telecommunications (DIET), ``Sapienza'' University of Rome, Via Eudossiana 18, 00184, Rome, Italy. Corresponding author: Simone Scardapane ({simone.scardapane@uniroma1.it})}}

\begin{document}


\maketitle

\begin{abstract}
\mbox{}\\
Deep neural networks are generally designed as a stack of differentiable layers, in which a prediction is obtained only after running the full stack. Recently, some contributions have proposed techniques to endow the networks with \textit{early exits}, allowing to obtain predictions at intermediate points of the stack. These multi-output networks have a number of advantages, including: (i) significant reductions of the inference time, (ii) reduced tendency to overfitting and vanishing gradients, and (iii) capability of being distributed over multi-tier computation platforms. In addition, they connect to the wider themes of biological plausibility and layered cognitive reasoning. In this paper, we provide a comprehensive introduction to this family of neural networks, by describing in a unified fashion the way these architectures can be designed, trained, and actually deployed in time-constrained scenarios. We also describe in-depth their application scenarios in 5G and Fog computing environments, as long as some of the open research questions connected to them.
\end{abstract}

\section{Introduction}
\label{sec:introduction}
\thispagestyle{plain}
The success of deep networks can be attributed in large part to their extreme modularity and compositionality, coupled with the power of automatic optimization routines such as stochastic gradient descent \cite{zhong2019automatic}. While a number of innovative components have been proposed recently (such as attention layers \cite{vaswani2017attention}, neural ODEs \cite{chen2018neural}, and graph modules \cite{schlichtkrull2018modeling}), the vast majority of deep networks is designed as a sequential stack of (differentiable) layers, trained by propagating the gradient from the final layer inwards.

Even if the optimization of very large stacks of layers can today be greatly improved with modern techniques such as residual connections \cite{zhang2018deep}, their implementation still brings forth a number of possible drawbacks. Firstly, very deep networks are hard to parallelize because of the gradient locking problem \cite{nokland2016direct} and the purely sequential nature of their information flow. Secondly, in the inference phase, these networks are complex to implement in resource-constrained or distributed scenarios \cite{park2019wireless,klaine2018distributed}. Thirdly, overfitting and vanishing gradient phenomena can still happen even with strong regularization, due the possibility of raw memorization intrinsic to these architectures \cite{jastrzkebski2017three}.

When discussing overfitting and model selection, these are generally considered properties of a given network applied to a full dataset. However, recently, a large number of contributions, e.g., \cite{lee2015deeply,wang2017deepgrowinglearning,marquez2018deep,belilovsky2018greedy,baccarelli2020optimized}, have shown that, even in very complex datasets such as ImageNet, the majority of patterns can be classified by resorting to smaller architectures. For example, \cite{leroux2017cascading} shows that the features extracted by a single convolutional layer already achieve a top-5 accuracy exceeding $30\%$ on ImageNet. Even more notably, \cite{wang2017idk,kaya2018shallow} show that predictions that would be correct with smaller architectures can become incorrect with progressively deeper architectures, a phenomenon the authors call \textit{over-thinking}.

In this paper, we explore a way to tackle with all these problems simultaneously, by endowing a given deep networks with multiple \textit{early exits} (auxiliary classifiers) departing from separate points in the architecture (see Fig. \ref{fig:early_exit}). Having one/two auxiliary classifiers is a common technique to simplify gradient propagation in deep networks, such as in the original Inception architecture \cite{szegedy2015going}. However, recently it was recognized that these multi-exit networks have a number of benefits, including the possibility of devising layered training strategies, improving the efficiency of the inference phase, regularizing for computational cost, just to name a few. Depending on which of these aspects takes priority on a certain application, in this paper we overview a number of techniques and ideas to design, train, and deploy this emerging class of neural network models.

\subsection{Contribution of this work}
\label{subsec:contribution_work}
While multi-exit networks have attracted significant interest lately, the related literature is widely fragmented, with similar ideas reappearing under broadly different names (e.g., cascaded networks \cite{marquez2018deep}, IDK networks \cite{wang2017idk}, deeply-supervised models \cite{lee2015deeply}, etc.). In this paper, we aim at providing a coherent and unified introduction to them, by highlighting the key challenges and opportunities related to their optimized design, training, and implementation.

To this end, this overview is broadly structured as follows. Section \ref{sec:related_works} briefly overviews three separate research lines that are closely connected to this paper. In Section \ref{sec:description_model} we describe the multi-exit neural network models. We also briefly introduce the main issues related to them, which are explored more in-depth in the following sections: designing and placing the auxiliary components (Section \ref{sec:placing_designing_early_exits}), training strategies (Section \ref{sec:training_the_architecture}), and multi-exit inference (Section \ref{sec:inference}). Then, in Section \ref{sec:additional_topics} we connect our discussion to some broader themes of interest, namely, biological plausibility of the training paradigms, distributed implementation on multi-tiered (e.g., networked) platforms, and the information bottleneck principle. We conclude by highlighting some open research challenges in Section \ref{sec:conclusion_open_research_questions}.

\section{Related works}
\label{sec:related_works}
Before moving to the main part of the overview, we briefly review three lines of works that are related to the topics we describe. This also allows us to properly define the domain of the paper.

\subsubsection*{Fast inference in neural networks}
Designing NN with fast execution times is a critical task, especially in resource-constrained applications (e.g., mobile and IoT). An early example is the work pursued by \cite{bucilua2006model}, where a very deep NN generates an endless stream of labelled examples for the adaptive training of a much smaller student network. More in general, several authors have considered lowering the implementation complexity by reducing the computational precision of the performed per-layer operations, either through compression (e.g., group sparsity), quantization, or the introduction of specific low-latency components \cite{hubara2016binarized,rastegari2016xnor}.

Multi-output neural networks are well suited for fast inference, but they exploit an orthogonal mechanism to the works described before. Instead of searching for a single, efficient network for processing all input patterns, they constrain the processing time to be small for the \textit{majority} of input patterns, through the introduction of early exits (see in particular Section \ref{sec:inference} later on). We avoid an overview of the relevant literature here, as it would be redundant with respect to the rest of the manuscript. We refer for this especially to Section \ref{sec:training_the_architecture}.

\subsubsection*{Distributed training of neural networks}
As we describe below, a second important motivation for multiple exits is to distribute model training and evaluation on a multi-tiered platform, such as in emerging Fog Computing (FC) applications. In fact, the general topic of the design of multi-tier FC platforms is receiving large attention in these last years, mainly due to the emerging areas of the Mobile Cloud, Pervasive Computing and Edge Computing \cite{othman2013survey,mach2017mobile}. These contributions focus on the offloading of (parts of) application programs from mobile devices to nearby FC servers through the exploitation of WiFi/Cellular-based communication technologies (see, for example, \cite{baccarelli2019ecomobifog} and references therein for an updated overview of this topic). NNs with multiple exits fit naturally into this paradigm, and we overview this more in-depth in Section \ref{subsec:distributed implementation}. A related topic is that of distributed training of NN architectures on networks of interrelated agents, e.g., \cite{scardapane2017framework}. As we discuss in Section \ref{subsec:distributed implementation} and Section \ref{subsec:layerwise_training}, some training approaches for multi-output networks naturally lend themselves to distributed implementations.

\subsubsection*{Layered training}
Having multiple early exits makes the NN a stage-wise classifier (see in particular Fig. \ref{fig:layerwise_training} and Section \ref{subsec:layerwise_training}). Layerwise training approaches have always been a fundamental area of research in neural networks. In fact, some of the earliest models predating back-propagation were iterative \cite{ivakhnenko1966cybernetic,belilovsky2018greedy}, and (unsupervised) layerwise strategies were instrumental in the training of some early very deep CNN models \cite{bengio2007greedy}. In this paper, we also survey layered optimization strategies in the context of multi-output neural networks. However, we underline that many recent lines of research on multi-layered training, such as boosting neural classifiers \cite{cortes2017adanet,huang2018learning}, kernel-based methods \cite{kulkarni2017layer}, progressive growing architectures \cite{karras2017progressive}, or clustering methods \cite{malach2018provably}, do not fit into our model.

\section{Description of the model}
\label{sec:description_model}

The discussion in Section \ref{sec:related_works} has framed several key concerns in the use of neural networks. We now turn to introducing formally the idea of multi-output networks, a design strategy that can be exploited to simplify training, reduce the inference cost, and distribute computation.

\subsection{Basic neural networks}
Consider a generic deep neural network $f(x)$, taking an input $x$ and providing a prediction $y$. The input $x$ can be as simple as a vector, or a more structured data type, such as an image, a sequence, a video, a graph, etc. Without loss of generality, we assume that $f$ is composed by the cascade of $L$ differentiable operators $f_1, \ldots, f_L$:
\begin{equation}
f(x) = (f_L \circ f_{L-1} \circ \ldots \circ f_1) (x) \,,
\label{eq:deep_network}
\end{equation}
where $\circ$ denotes function composition such that $(f_i \triangleq f_j)(\cdot) = f_i(f_j(\cdot))$. Examples of operations that can act as components in \eqref{eq:deep_network} include convolutional layers for images \cite{goodfellow2016deep}, batch normalization, self-attention \cite{vaswani2017attention}, and many others.\footnote{A single $f_i$ can also be a sequence of operations. We do not make any distinction in the paper, and we use the terms \textit{block} and \textit{layer} indistinguishably to refer to a single $f_i$.} We denote by $h_i$ the output of the $i$th operation, so that $h_i \triangleq f_i(h_{i-1})$, and $h_L = y$ is the final output of the network. It is common to refer to $h_i$ as the $i$th \textit{embedding} generated by the network for the input $x$. Finally, $\theta_i$ will denote the set of adaptable parameters of the $i$th layer, which can eventually be empty (e.g., in the case of dropout).

Given a set of examples $\mathcal{D} = \left\{(x_n, y_n)\right\}_{n=1}^N$ and a loss function $l(\cdot, \cdot)$ penalizing the wrong predictions of the network, the parameters are tuned by performing stochastic gradient descent (SGD) on the expected loss:
\begin{equation}
f^* = \arg\min_\theta \sum_{n=1}^N l\left( y_n, f(x_n) \right) \,,
\label{eq:final_loss}
\end{equation}
where $\theta = \bigcup_{i=1}^L \theta_i$ is the union of all trainable parameters. In this basic setup, all $L$ layers must be evaluated before obtaining a prediction and this full processing, as we elaborated on in the introduction, might not be necessary and even downright harmful in terms of overfitting and energy consumption. Next, we describe a model that attempts to address these drawbacks by adding intermediate prediction steps along the network stack.

\subsection{Neural networks with early exits}
\label{subsec:a_neural_network_with_early_exits}

\begin{figure}
	\centering
	\includegraphics[width=0.6\columnwidth]{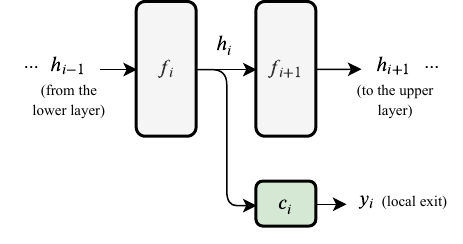}
	\caption{Graphical depiction of a generic early exit in neural network architectures. In green, we show the auxiliary predictor.}
	\label{fig:early_exit}
\end{figure}

In order to describe this extended architecture, we begin by selecting a set of early exit indices $\mathcal{C} \subseteq \left\{ 1, \ldots, L-1\right \}$ corresponding to `interesting' mid-points in the architecture. We will have more to say on how to properly select $\mathcal{C}$ in the following sections of the paper. At each mid-point $i \in \mathcal{C}$ that was selected, we feed the intermediate embedding $h_i$ to an auxiliary classifier/regressor $c_i$ to obtain an intermediate prediction $y_i$:
\begin{equation}
y_i = c_i(h_i) \,,
\end{equation}
where $c_i$ is another (smaller) auxiliary neural network in the form \eqref{eq:deep_network}, typically consisting of only one/two layers. We call $f$ the \textit{backbone} network, $c_i$ the $i$th auxiliary classifier/regressor, and $y_i$ the $i$th \textit{early-exit} from the backbone. This setup is depicted graphically in Fig. \ref{fig:early_exit}. As a result, by running the full network with all early exits we obtain a sequence of predictions $y_i$, $i \in \mathcal{C}$, each potentially more refined and accurate than the previous one. However, in order to make the specification of this model useful, a number of questions must be answered, forming the basis for the rest of this overview. We briefly summarize the key points below. Note that some design choices are mutually dependent (e.g., the training strategy can influence the criterion for placing the auxiliary classifiers). However, for clarity, we try to keep them conceptually separated and mention inter-dependencies when needed in the corresponding sections.

\subsubsection{Selecting where and how to early-exit}
Firstly, one has to select the design of the auxiliary models, the number of early exits, and their placements over the backbone network. If the multi-exit architecture is used only to counter overfitting, then a small number of early exits is generally sufficient. For example, the original Inception model described in \cite{szegedy2015going} has two early exits, placed roughly at $1/3$ and $2/3$ of the backbone network. More in general, however, placing a set of early exits to satisfy energy or efficiency constraints is a combinatorial problem. This problem is formally described in Section \ref{sec:placing_designing_early_exits}.

\subsubsection{Training the network}
Secondly, a proper training mechanism must be devised. Here, we partition the possible training approaches in three broad classes:
\begin{enumerate}
	\item The overall architecture can be trained \textit{jointly} by defining a single optimization problem that embraces all intermediate exits. This can be done by a combination of per-classifier losses \cite{lee2015deeply}, or by first combining the outputs and then building a single loss atop this combination \cite{scardapane2020differentiable}.
	\item We can adopt a \textit{layer-wise} approach, where at each iteration we train a single auxiliary classifier together with the backbone layers preceding it, and, then, we freeze them after training \cite{hettinger2017forward}.
	\item We can first train the backbone network, and then separately train the auxiliary classifiers on top of it \cite{venkataramani2015scalable}. This can be interpreted as a very primitive form of knowledge distillation \cite{hinton2015distilling}.
\end{enumerate}
We describe these three classes, along with some design considerations, in Section \ref{sec:training_the_architecture}.

\subsubsection{Exploiting multiple early exits in the inference phase}
Thirdly, once the network is trained, a suitable inference mechanism must be chosen. Once again, in the simplest case, auxiliary classifiers are used only to simplify gradient flow, and in this case their intermediate exits can be discarded \cite{szegedy2015going}. Alternatively, the different predictions can be combined to obtain an ensemble almost for free \cite{scardapane2020differentiable}. In the most interesting case, however, one wants to exploit the different exits to process each input as efficiently as possible, by selecting the first auxiliary classifier that has a reasonable chance to provide the correct prediction, exploiting the full network only for cases that are extremely hard to predict \cite{zhou2019edge}. In this case, it is necessary to select (and possibly train) a criterion to decide whether to exit the network or whether to continue processing over the next block. We consider this task in Section \ref{sec:inference}.

\subsubsection{Formal properties of multi-exit networks}
Finally, we can consider the properties induced by the separate exits, in terms of, e.g., non-convexity of the optimization problem \cite{zhang2018deep}, convergence and generalization gaps of the different training approaches, guaranteed accuracy of the intermediate predictions, and so on.  While the focus of this survey is more application-oriented, we still overview some of the most interesting results on these formal topics later on.

\section{Placement and optimized design of the early exits}
\label{sec:placing_designing_early_exits}
To begin with, we need to devise an optimized strategy for designing and implementing the auxiliary classifiers $c_i$.\footnote{We use the term classifier for simplicity, but everything extends to regression and other supervised problems.} For fully-connected layers, the vast majority of works consider either simple linear classifiers \cite{teerapittayanon2016branchynet}, or small neural networks with one/two hidden layers \cite{kaya2018shallow}. For convolutional layers, this is slightly more complex because early layers in most CNNs have a very high dimensionality, which would result in extremely high-dimensional classifiers \cite{kaya2018shallow}. A simple solution in this case is to consider very strong dimensionality reduction procedures before the classification step, in the form of global average pooling or pooling with very large windows.

A similar consideration has to be taken into account when designing the set of indexes $\mathcal{C}$ for the placements of the early exits. Properly selecting this set is especially important for the case in which early exits must be used to speed-up the inference phase. A very simple technique is to estimate the relative computational cost of each layer, and place the early exits to some given percentiles of the network' total cost \cite{kaya2018shallow}.

More in general, let us suppose that $\gamma_i$ denotes the computational cost of running the network up to early exit $i$ (as measured, e.g., in total number of operations to be performed). Similarly, denote by $I_i$ the fraction of data that exits at the $i$th early exit (techniques to decide when to stop are discussed later on in Section \ref{sec:inference}). Focusing only on early exit $i$, using the auxiliary classifier improves the efficiency of the network whenever the following relation is met \cite{panda2017energy}:
\begin{equation}
\left(\gamma_{i+1} - \gamma_i\right)\left(I_i - I_{i+1}\right) > \gamma_i I_{i+1} \,.
\label{eq:efficiency_test}
\end{equation}
Because the number of possible placements of $M$ early exits in a network grows exponentially in the network's depth $L$, the optimal placement according to \eqref{eq:efficiency_test} can only be evaluated for small networks. 

However, possibly sub-optimal greedy algorithms have been proposed to evaluate the placement from the original of the network up to the final layer \cite{panda2017energy,baccarelli2020optimized}. The algorithm in \cite{baccarelli2020optimized}, in particular, allows to define a further hyper-parameter $\text{TH} \in \left[0, 1\right]$ to fine-tune the final validation accuracy. To use the same description as in \cite{scardapane2020differentiable}, we overload our previous notation and define $\gamma_{f_i}$ to be the computational cost of $f_i$ alone, and similarly for $\gamma_{c_i}$. In addition, we define a compression ratio $\text{cm}_i = \frac{I_{i+1}}{I_i}$. Then, the greedy approach in \cite{baccarelli2020optimized} keeps an early exit whenever the following relation at layer $i$ is met:
\begin{align}
\left(\text{TH} - \text{cm}_i\right)&\gamma_{f_{i+1}} -  \left(1 - \text{cm}_i\right)\gamma_{c_i} - \nonumber\\ & \left(1 - \text{TH}\right)\gamma_{f_i} \ge 0 \,.
\end{align}
An alternative approach, described in \cite{bolukbasi2017adaptive}, is to evaluate each early exit considering both constant auxiliary classifiers and adaptive auxiliary classifiers. Whenever the former performs better in terms of training / efficiency trade-off, we remove them from our set. General hyper-parameter optimization procedures \cite{stamoulis2018designing} can also be used to fine-tune the entire architecture, including the eventual placement of some early exits.

\section{Training neural networks with early exits}
\label{sec:training_the_architecture}
In this section, we consider the second problem introduced in Section \ref{sec:description_model}, namely, training the architecture. We describe several variants that have been proposed in the literature, and we highlight some of their formal properties. Note that, once the set $\mathcal{C}$ of the early exit placements has been designed, we can always assume that there is a one-to-one correspondence between layers and auxiliary classifiers, since in-between layers can be merged into a single layer. Because this simplifies our notation, we will make this assumption from here on. This section is organized as follows: we start with joint training strategies in Sections \ref{subsec:joint_training} and \ref{subsec:joint_training_on_combined_output}, then we describe layer-wise strategies in Section \ref{subsec:layerwise_training}, and we conclude with other variants in Section \ref{subsec:additional_training_approaches}.

\subsection{Joint training approaches}
\label{subsec:joint_training}

\begin{figure*}
	\centering
	\includegraphics[width=0.85\columnwidth]{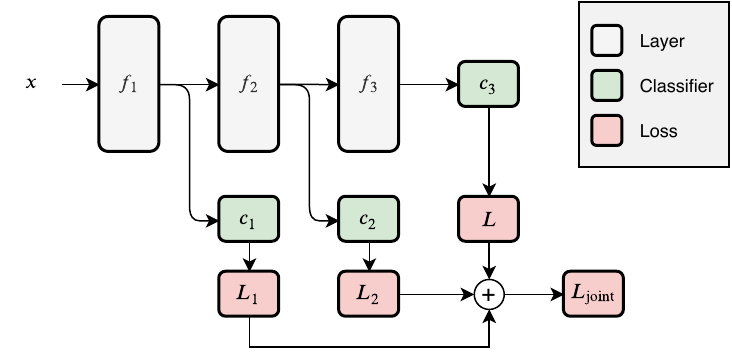}
	\caption{Graphical description of the joint training approach. Colors are described in the legend.}
	\label{fig:joint_training}
\end{figure*}

In the simplest setup, we can associate a loss $L_i$ to each intermediate classifier $c_i$, and perform SGD on a suitable combination of all losses. We call this \textit{joint training} (JT) or, equivalently, deeply-supervised \cite{lee2015deeply}. Specifically, we train the model on:
\begin{equation}
f^* = \arg\min\left\{ L + \sum_{i=1}^{L-1} \alpha_i L_i \right\} \,,
\end{equation}
where $L$ (called the \textit{overall loss} in \cite{lee2015deeply}) is the final loss defined in \eqref{eq:final_loss}, $L_i$ is an auxiliary loss (called \textit{companion loss} in \cite{lee2015deeply}) defined as:
\begin{equation}
L_i \triangleq \sum_{n=1}^N l\left(y_n, c_i(x_n)\right) \,,
\label{eq:joint_loss}
\end{equation}
and $\alpha_i$ weights the contribution of the $i$th auxiliary classifier. We will discuss the selection of these weights later on. In general, if the auxiliary classifiers are only used to improve performance, it is customary to weight earlier classifiers less. For example, the Inception model \cite{szegedy2015going} propose to set $\alpha_i = 0.3$. A schematic description of the joint training method is provided in Fig. \ref{fig:joint_training}.

Apart from Inception, similar approaches have been successfully applied to, among others, face recognition \cite{sun2015deeply}, object detection \cite{cai2016unified}, image segmentation \cite{chen2016attention}, pose estimation \cite{insafutdinov2016deepercut}, saliency detection \cite{liu2016dhsnet}, metric learning \cite{yuan2017hard}, visual attention prediction \cite{wang2017deep}, and super resolution \cite{lai2017deep,tong2017image}. JT approaches are also popular in the case of recurrent neural networks for sequence classification / regression, in which intermediate predictions can be obtained for free by considering the outputs of the network at each time-step \cite{lipton2015learning}.

\subsubsection{On the convergence property of the JT approach}
Interestingly, it can be shown that even a relatively simple setup as \eqref{eq:joint_loss} can improve the convergence of the resulting network with early exits. \cite{lee2015deeply} analyzes the case where the loss is selected as the squared hinge loss (although the proof can be easily extended to other losses, such as the cross-entropy), and each loss is augmented by an $\ell_2$ regularization term. In this case, they prove in \cite{lee2015deeply} the following result.
\begin{theorem}[Theorem 1, \cite{lee2015deeply}]
	Assume that $\sum_{i=1}^{L-1} L_i$ is $\lambda$-strongly convex\footnote{A function F is $\lambda$-strongly convex if, for any $a$, $b$, we have that $F(a) \ge F(b) + \nabla^T F(b)(a-b) + \frac{\lambda}{2}\lVert a-b \rVert^2$.} close to the optimal $\theta^*$. Let us denote by $T$ the number of iterations of SGD, by $\eta_t = 1/t$ the decaying step size at iteration $t$, by $\theta_T^{\text{standard}}$ the solution obtained by SGD at iteration $T$ when optimizing \eqref{eq:final_loss}, and by $\theta_T^{\text{joint}}$ the one obtained by optimizing \eqref{eq:joint_loss}. Then, optimizing \eqref{eq:joint_loss} compared to \eqref{eq:final_loss} provides a relative speed-up in convergence of $\frac{\mathbb{E}\left[ \lVert \theta^* - \theta_T^{\text{standard}} \right \rVert^2 ]}{\mathbb{E}\left[ \lVert \theta^* - \theta_T^{\text{joint}} \rVert^2 \right]} = \Theta(\exp\{\ln(T)\lambda)\})$.
\end{theorem}
The authors in \cite{lee2015deeply} argue that, in practice, the Lipschitz constant $\lambda$ of the companion loss is large (compared to the Lipschitz constant of the original loss), providing a justification for the good empirical performance of the JT approach. However, analyzing and proving the relationship between the inner (local) and standard optimization problems remain open questions. Also, we note that this strategy could benefit from a more efficient sampling of patterns at the different exits, although this aspect has not received much attention in the literature.

\subsection{Joint training on a combined output}
\label{subsec:joint_training_on_combined_output}
A variant of the JT approach is described in \cite{kim2016deeply}. Instead of merging the auxiliary losses, we can adaptively merge the \textit{predictions} of the network as:
\begin{equation}
\hat{y} = \alpha f(x) + \sum_{i=1}^{L-1} \alpha_i c_i(x) \,,
\label{eq:sum_output}
\end{equation}
where the weights $\alpha$ and $\alpha_i$ can be trained simultaneously with $\theta$. Several variants can be devised where, for example, weights are constrained to be positive, or they are processed through a softmax function to arrive at a convex combination.

More in general, we can devise a range of options to adaptively combine the different auxiliary predictions. For example, \cite{teerapittayanon2017distributed} analyze max-pooling and average-pooling to combine the different predictions with a fixed (non-trainable) operation. Alternatively, \cite{scardapane2020differentiable} endows each early exit with an additional binary classifier $g_i(x) \in \left[0, 1\right]$ to decide how much the current prediction should be trusted. In this case, the global output can be defined recursively as:
\begin{equation}
\hat{c}_i(x) = g_i(x)\,c_i(x) + (1-g_i(x))\,\hat{c}_{i+1}(x) \,,
\label{eq:recursive_output}
\end{equation}
where the base case is the final prediction $\hat{c}_{L}(x) = f(x)$. The combination process can also be implemented through an additional network component (e.g., an LSTM or recurrent network), providing a further degree of freedom to the system. This last setup is related to the use of boosting methods to incrementally build the network's architecture \cite{cortes2017adanet,nitanda2018functional}.

\subsection{Layer-wise training}
\label{subsec:layerwise_training}

\begin{figure*}[!th]
	\centering\hfill{}
	\subfigure[Layerwise training]{\includegraphics[width=0.48\columnwidth]{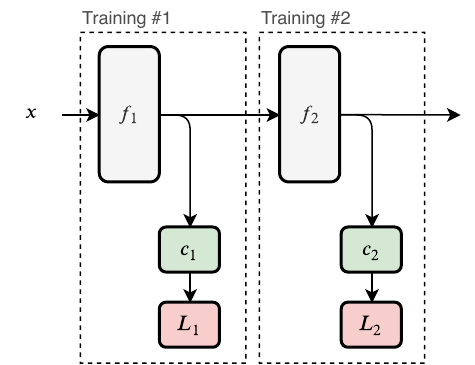}\label{fig:layerwise_training}}
	\subfigure[Separate training]{\includegraphics[width=0.48\columnwidth]{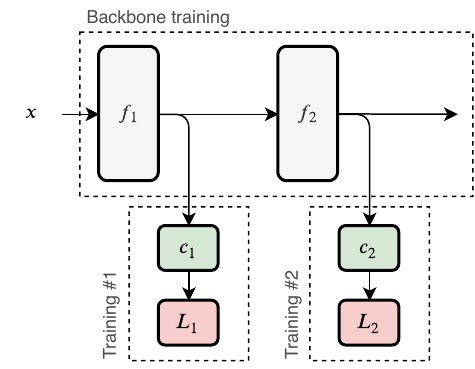}\label{fig:separate_training}}\hfill{}
	\caption{Graphical depiction of the (a) layerwise training approach and (b) separate training approach. Colors are the same used in Fig. \ref{fig:joint_training}.}
	\label{fig:alternative_training_strategies}
\end{figure*}

The insertion of auxiliary local classifiers opens up interesting alternatives to the standard joint training described above. In particular, we can devise a \textit{layer-wise training} (LT) strategy by separately optimizing each layer in the architecture. Note that supervised and unsupervised LT methods were explored repeatedly in the initial attempts at building deeper architectures \cite{bengio2007greedy,larochelle2009exploring}. Only recently, however, have they become a successful tool for training them, e.g., through boosting \cite{cortes2017adanet,huang2018learning}, clustering \cite{malach2018provably}, and kernel similarity \cite{kulkarni2017layer}. 

In the context of multi-exit networks as described in this paper, \cite{hettinger2017forward} introduced the general idea under the name `forward thinking', while \cite{marquez2018deep} calls it `deep cascade learning'. We depict it visually in Fig. \ref{fig:layerwise_training}.

According to Fig. \ref{fig:layerwise_training}, the LT approach proceeds as follows:
\begin{enumerate}
	\item Start by training the first auxiliary classifier $c_1(f_1(x))$, depicted as \textit{Training \#1} in Fig. \ref{fig:layerwise_training}. Considering the notation of Section \ref{subsec:joint_training}, this is equivalent to minimize $L_1$ individually.
	\item Freeze the previous layer, by replacing all inputs $x_i$ in the training set with the corresponding embedding $f_1(x_i)$.
	\item Train the second layer $c_2(f_2(\cdot))$, by solving $L_2$ only with respect to the weights $\theta_2$.
	\item Repeat steps (2)-(3) for as many layers as needed.
\end{enumerate}
The advantage of this method is that, since each layer is relatively small (depending on the size of each $f_i$ and $c_i$), regularization problems such as vanishing and exploding gradients should be less prominent under this training approach \cite{hettinger2017forward}. An extension to semi-supervised problems with growing networks is presented in \cite{wang2017deepgrowinglearning}.

\subsubsection{On the formal properties of the LT approach}
A preliminary formal analysis of the LT approach is provided in \cite{belilovsky2018shallow}, and later extended in \cite{belilovsky2018greedy}. Roughly speaking, their works show that, if each layer-wise optimization process incurs an error that is upper bounded by $\varepsilon$, the final predictions given by the network will have an error with a similar upper bound given by $\mathcal{O}(L^2\varepsilon)$. They also show, empirically and through an informal analysis, that the approach will naturally lead to increasing the linear separability of the data for each layer that is added. Notably, \cite{belilovsky2018greedy} is the first work to show that LT can scale to large-scale problems, such as image classification on ImageNet. Additional formal considerations are provided in \cite{baccarelli2020optimized} showing that, under suitable considerations on the blocks composing the architecture, there always exists a solution in which adding a new layer does not worsen the empirical error of the model.

\subsection{Further training approaches}
\label{subsec:additional_training_approaches}

Besides joint training (Sections \ref{subsec:joint_training} and \ref{subsec:joint_training_on_combined_output}) and layer-wise training (Section \ref{subsec:layerwise_training}), the flexibility of the architectures analyzed in this paper lend themselves to a variety of alternative training approaches, which we briefly describe in the sequel. First of all, several authors \cite{venkataramani2015scalable,panda2016conditional,bolukbasi2017adaptive,kaya2018shallow} have considered a variation of LT, in which the backbone network is trained separately from each of the auxiliary classifier. This is depicted visually in Fig. \ref{fig:separate_training}. This kind of approach is especially suitable for describing and analyzing situations in which the backbone network is provided beforehand, and we desire to turn it into a multi-exit architecture. It also allows for easy experimentation on the selection of the auxiliary classifiers, which can be trained and removed quickly according to the rules-of-thumb introduced in Section \ref{sec:placing_designing_early_exits}. Finally, it allows for the use of non-neural auxiliary classifiers \cite{venkataramani2015scalable}, such as using decision trees or support vector machines for the intermediate predictions.

\begin{figure}
	\centering
	\includegraphics[width=0.5\columnwidth]{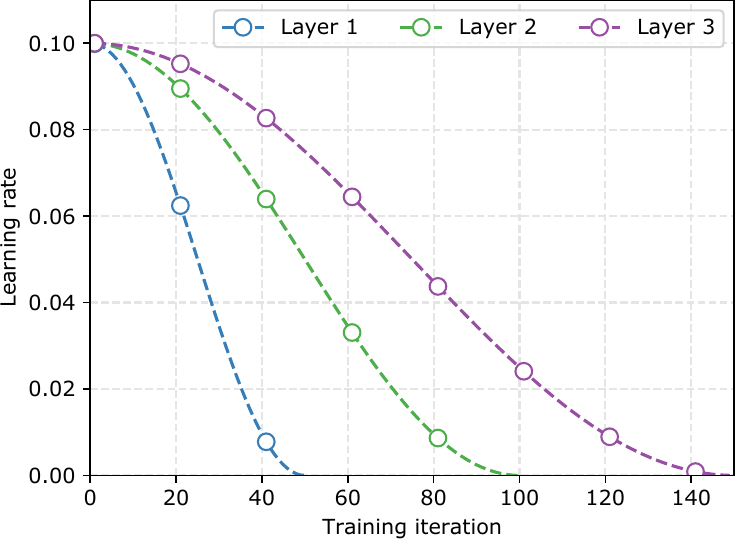}
	\caption{Example of progressively freezing out layers as in \cite{brock2017freezeout}.}
	\label{fig:cosine_annealing}
\end{figure}

Some strategies mix different approaches. For example, freezeout \cite{brock2017freezeout} is a variation of the JT approach, in which layers are iteratively `freezed' by annealing their learning rates to zero, providing a trade-off between the accuracy of training everything jointly with a partial speed-up similar to a layer-wise training. Specifically, denote by $\eta_i(t)$ the learning rate of layer $i$ at iteration $t$. Given a budget of training iterations $T$, we select a set of freezing points $t_i$ equispaced in $[0, T]$. Then, the learning rate of layer $i$ for $t < t_i$ is given by:
\begin{equation}
\eta_i(t) = 0.5 \, \eta_i(0) \left( 1 + \cos\left( \frac{\pi t}{t_i} \right)  \right) \,,
\label{eq:cosine_annealing}
\end{equation}
while $\eta_i(t) = 0$ for $t > t_i$. An example with $L=3$, $T=150$, and $\eta_i(0) = 0.1$ is shown in Fig. \ref{fig:cosine_annealing}.

To conclude this section, we also mention that some authors have considered reinforcement learning strategies to train the overall architecture \cite{guan2017energy}.

\section{Inference in multi-exit networks}
\label{sec:inference}
We now focus on the topic of performing inference in architectures with multiple auxiliary classifiers. Note that, in some cases, this is trivial. For example, in architectures such as Inception \cite{szegedy2015going}, where the auxiliary classifiers are only used for improving the training phase, one can discard them and use $f(x)$ as the single final exit. In other cases, such as the aggregated output from \eqref{eq:sum_output} or \eqref{eq:recursive_output}, the network automatically provides a joint prediction that merges all the intermediate ones.

In the most interesting case, however, we can assume that every input $x$ has an optimal depth corresponding to one specific prediction $c_i(x)$. In this case, we prefer to process each input up to the corresponding exit and, then, stop the forward propagation. In fact, similar schemes have been shown to provide significant improvements in inference time \cite{teerapittayanon2016branchynet,teerapittayanon2017distributed}, and they can also prevent overfitting. In particular, \cite{kaya2018shallow} describes a phenomenon it calls \textit{over-thinking}, in which a prediction $c_i(x)$ on the test set is correct, but later predictions turn out to be wrong.

For classification problems, a common setup in this case is to consider the confidence of the network on its own prediction, and exploit this to decide whether to early-exit the architecture. According to this setup, denote by $c_{it}(x)$ the prediction of the $i$th classifier on the $t$th class, and by $C$ the total number of classes. The normalized entropy of the prediction is:
\begin{equation}
H\left[ c_i(x) \right] \triangleq \frac{1}{\log\left(C\right)}\sum_t c_{it}(x)\log\left( c_{it}(x) \right) \,.
\end{equation}
The value $H\left[ c_i(x) \right]$ is always normalized between $0$ and $1$. By comparing $H\left[ c_i(x) \right]$ against a user-selected threshold $\beta_i$, we can make the final decision on whether to exit the architecture. This strategy was popularized by BranchyNet 
\cite{teerapittayanon2016branchynet,teerapittayanon2017distributed}. The drawback of this approach is that a separate threshold must be selected for each auxiliary classifier, in order to ensure that the average accuracy of the model does not degrade. \cite{wang2017idk} describes other alternatives, including using the most-confident prediction instead of the entropy, or learning the confidence score on top of $c_i(x)$ with a separate trainable function $g_i(c_i(x))$. The gating approach described in \eqref{eq:recursive_output} is a specialization of this idea.

Alternatively, \cite{baccarelli2020optimized} introduces an iterative approach to fine-tune a single threshold given a desired accuracy on the level, based on a gradient-like approach exploiting the error between the current level of accuracy and the desired one. The strategy is also proved in \cite{baccarelli2020optimized} to be convergent.

\subsection{Regularizing for computational cost}
\label{subsec:regularizing_cost}
One interesting side benefit arising from the utilization of early exits  we study here is that we can regularize the resulting networks for \textit{efficiency}. Searching for efficient architectures is a wide problem in the literature on neural model selection \cite{pham2018efficient}. Having multiple auxiliary classifiers allows for an unprecedented opportunity of optimizing the accuracy/efficiency trade-off for each input \cite{scardapane2020differentiable}.

To this end, denote by $\delta_{ij}$ an indicator function describing whether the $i$th training pattern was processed by the $j$th auxiliary classifier. In addition, denote by $\epsilon_j$ the cost of executing the $j$th auxiliary classifier as compared to the $(j-1)$th one. Cost here is a generic quantity, that we can define to simply consider the number of operations executed by the auxiliary classifier (in this case $\epsilon_j = \gamma_{c_j}$ as defined in Section \ref{sec:placing_designing_early_exits}), or a more abstract definition, which can include the communication costs in a distributed environment \cite{zhou2019edge}. The average cost of $f$ is, then, defined as follows:
\begin{equation}
C[f] \triangleq \sum_{i=1}^n \sum_{j=1}^L \delta_{ij} \cdot \epsilon_j \,.
\label{eq:computational_cost}
\end{equation}
Relation \eqref{eq:computational_cost} can now be used in any procedure to select the hyper-parameters of the model, in order to explicitly trade-off the average inference cost of the neural network. \cite{wang2017idk} calls \eqref{eq:computational_cost} the `cascaded computational cost'.

In addition, supposing that the exit strategy $g_i(x)$ is differentiable (such as the entropy measure described previously), similarly to \eqref{eq:recursive_output}, we can define a soft-approximation to \eqref{eq:computational_cost} as \cite{scardapane2020differentiable}:
\begin{equation}
C[f] = \sum_{i=1}^n \sum_{j=1}^L p_{ij}\epsilon_j \,,
\end{equation}
where:
\begin{equation}
p_{ij} = g_j(x_i) \cdot \prod_{l=1}^{j-1} (1-g_l(x_i)) \,.
\end{equation}
Since the above expression is differentiable, we can use it as an alternative to the classical $\ell_p$ regularization to provide an explicit trade-off concerning accuracy and efficiency of the network, as described by the average computation cost incurred for a single example.

In the case of a pre-trained backbone network, \cite{bolukbasi2017adaptive} explores training separately each auxiliary classifier with a constraint on computational time. The global problem is, then, reduced to a sequence of weighted binary classification problems, where the weights depend on the costs. This is extended in \cite{stamoulis2018designing}, where the entire network design (including multiple hyper-parameters) is optimized through a Bayesian optimization procedure.

\section{Additional topics}
\label{sec:additional_topics}

In this section we detail three interesting applicative fields for the multi-output architecture described in the paper. We start with the topic of biological plausibility of deep networks and their training algorithms in Section \ref{subsec:biological_plausibility}, move on to a distributed implementation over multi-tiered or distributed environments in Section \ref{subsec:distributed implementation}, and conclude with an overview of layer-wise training in the context of the information bottleneck principle in Section \ref{subsec:the_information_bottleneck_principle}.

\subsection{Biological plausibility of multi-exit networks}
\label{subsec:biological_plausibility}
Multi-exit architectures in their LT formulation provide an interesting perspective on biological plausibility of neural network training \cite{mostafa2018deep}. In fact, backpropagation in classical neural networks is generally considered to be biologically implausible for multiple reasons \cite{betti2018backpropagation}, i.e.,: (i) the need to store the outputs for each intermediate layer $f_i$, to re-use it in the backward pass; (ii) the symmetry of weight matrices between forward and backward passes; and (iii) the perfect synchronization required by the process. Point (ii) has been explored in-depth in the literature on feedback alignment (and similar) \cite{lillicrap2016random,nokland2016direct,baldi2018learning}, in which it is shown that the transpose of the weight matrix can be replaced by a (fixed) random matrix during the backward pass while still allowing for convergence of the algorithm. This approach, however, still requires a global error to be back-propagated from the final output layer, thus failing to satisfy points (i) and (iii).

\cite{mostafa2018deep} explores a variation of the LT approach, in which each auxiliary classifier is set as $c_i(x) = Mf_i(x)$, where $M$ is a random matrix, which is eventually replaced by a separate random matrix $K$ (of the same size) during the backward pass. This approach can still obtain good results in training (albeit not at the level of a classical LT training), while being implementable through highly compact update equations involving only local error terms, thus drastically improving the overall biological plausibility of the model.

How to extend this framework to improve the classification results closer to the state-of-the-art while not sacrificing the biological plausibility remains an open research question. For example, \cite{nokland2019training} further replaces the true label at each auxiliary exit with a randomly projected version, while at the same time increasing the supervised loss with a separate one that forces the embeddings of the network to be similar whenever the corresponding projected outputs are similar. In \cite{belilovsky2019decoupled}, the authors investigate this approach for parallel optimization of the network, where a set of intermediate buffers is added to ensure the possibility of decoupling a layer from the previous ones. Based on the literature on target propagation \cite{lee2015difference}, the authors in \cite{kao2019associated} propose a further extension to decouple the different updates, in which the targets for the different early exits are computed adaptively from the original output $y_i$ by an additional set of autoencoder networks.

\subsection{Distributed implementation of multi-exit networks}
\label{subsec:distributed implementation}

Many recent computation frameworks, ranging from 5G cellular networks to IoT sensors and fog computing, are built upon the idea of multi-tier computations, in which decisions can be taken both at the edge of the networks or delegated to other, more powerful decision centers that incur some communication latency \cite{park2019wireless,zhou2019edge}. In these works, it is common to design systems in which separate neural networks, of different complexities, can run on the different tiers or on different computational units \cite{lo2017dynamic,nan2017adaptive}.

\begin{figure}
	\centering
	\includegraphics[width=0.55\columnwidth]{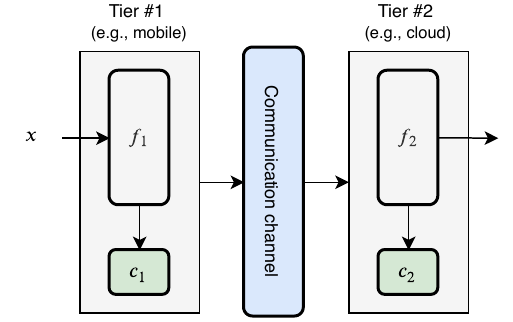}
	\caption{Distributed implementation of a multi-exit neural network on separate tiers of an underlying distributed computing platform.}
	\label{fig:distributed_implementation}
\end{figure}

Multi-exit neural networks fit naturally inside these emerging classes of technologies, by distributing different auxiliary classifiers over the different tiers, as depicted in Fig. \ref{fig:distributed_implementation}. In this case, an estimation of the cost of communicating across the channel can be included both in the design of the early exits, as described in Section \ref{sec:placing_designing_early_exits}, or in the model selection / regularization phase, as described in Section \ref{subsec:regularizing_cost} \cite{leroux2017cascading}.

An interesting question concerns the additional implementation of a distributed training phase \cite{scardapane2017framework}, which is necessary whenever the separate tiers have to adapt to new data without the possibility of communicating everything to a centralized controller. LT methods lend themselves naturally to an iterative implementation, in which every tier trains its own model before sending the new set of input features to the next tier. Alternatively, more sophisticated distributed optimization strategies, possibly taking into account also communication constraints, are an open interesting research area \cite{barbarossa2014communicating,baccarelli2019ecomobifog}.

\subsection{The information bottleneck principle}
\label{subsec:the_information_bottleneck_principle}
The last application context that we consider is that of the information bottleneck (IB) principle \cite{amjad2019learning}. The IB was proposed as a general principle to explain the expressive power of deep networks, relying on information theoretic principles. Interestingly, one possibility for the actual implementation and test of the IB principle results in a different training scenario for multi-exit neural networks \cite{elad2018effectiveness}.

Slightly overloading our notation, denote by $X$ a random variable describing the inputs to the network, by $Y$ a second random variable describing the corresponding targets, and by $F_i$ a third random variable describing the embeddings generated by layer $f_i$. In addition, denote by $I(\cdot, \cdot)$ the mutual information between two random variables. Roughly, the IB principle states that neural networks work because their intermediate representations tend to simultaneously maximize their mutual information with the target, while at the same time minimizing their mutual information with the input. Stated differently, they tend to compress information as much as possible while keeping the most predictive bits of information. Stated more formally, the IB principle states that intermediate representations tend to minimize:
\begin{equation}
\mathcal{L}_i = I\left(Y, F_i\right) - \beta \, I\left(X, F_i\right) \,,
\label{eq:ib_principle}
\end{equation}
where $\beta$ is a trade-off parameter.\footnote{Note that, in practice, the term in \eqref{eq:ib_principle} is infinite for deterministic networks with continuous inputs. However, it can be replaced with a noisy approximation \cite{saxe2019information}.} While the IB principle has received ample empirical validation \cite{saxe2019information,shwartz2017opening}, it is common to evaluate and/or apply it only to a \textit{single} layer of the network.

\cite{elad2018effectiveness} proposes to exploit the IB principle as a training criterion for neural networks, wherein the network is trained according to a LT approach, substituting the local loss term $L_i$ in \eqref{eq:joint_loss} with an approximation of \eqref{eq:ib_principle}. Interestingly, this approximation requires the addition of multiple auxiliary classifiers to the network, acting as discriminators between embeddings corresponding to separate classes. Alternative supervised and unsupervised training strategies have also been proposed in a number of papers \cite{elad2019direct,lowe2019putting}.

\section{Open research challenges}
\label{sec:conclusion_open_research_questions}

In this paper, we provided a self-contained, unified introduction to the design, implementation, and training of neural network with multiple early exits. As we argued, these models have a number of favourable properties, including a higher resilience to gradient problems, as long as a large flexibility to more advanced training / inference strategies. They also provide a viable entry point to the study of more biologically plausible alternatives to classical back-propagation, and they are more naturally suited to an implementation in multi-tiered computational frameworks such as the emerging fog computing one \cite{baccarelli2017fog}.

Apart from the multiple minor points discussed in the paper, we identify three main areas where research is needed in this context.

\subsubsection*{Challenge 1: How powerful are multi-output neural networks?}
The formal properties of these more general network architectures are only recently starting to be investigated, as we described in this paper, leaving ample margins to, e.g., quantify the relative generalization gap of each component. In addition, while these techniques have been applied multiple times to CNNs and, in smaller measure, to RNNs, their usefulness remains to be tested on more modern neural layers, such as graph convolutional networks, and transformer architectures. Finally, how to properly consider both architectural constraints, accuracy, and communication costs when implementing the networks in a distributed environment is still an open question.

\subsubsection*{Challenge 2: Full integration in FC/IoT environments}
In principle, multi-output NNs promise to be a perfect match for distributed, multi-tiered systems given by the convergence of several recent paradigms, most notably Fog Computing \cite{baccarelli2019ecomobifog}. A number of challenges, however, need to be solved in order to see this convergence in practice, including elastic, asynchronous training strategies, optimal placement of the early exits, efficient inference and communication costs, and so on. Several of these challenges are described and reviewed in \cite{baccarelli2020optimized}.

\subsubsection*{Challenge 3: What comes after having multiple exits?}

More in general, deep learning has risen to prominence thanks to the extreme modularity and flexibility of automatic differentiation frameworks combined with modern programming languages. We believe that architectures such as those explored in this paper represent a natural next step in moving away from purely feedforward implementation to more complex, multi-exit, multi-resolution ones. In this sense, an immediate extension of the ideas presented here is to consider multiple networks, conditionally connected on one another, performing inferences or routing information to one another \cite{bolukbasi2017adaptive}.

\bibliographystyle{spmpsci}
\bibliography{biblio}

\end{document}